\documentclass[letterpaper, 10 pt, journal, twoside]{ieeetran}

\IEEEoverridecommandlockouts                          

\usepackage{graphicx}
\usepackage[dvipsnames]{xcolor}

\usepackage{amsmath}
\usepackage{amssymb}
\usepackage{amsfonts}
\usepackage{mathtools}
\usepackage{mathrsfs}

\usepackage{multirow}
\usepackage{booktabs, tabularx, makecell, colortbl, array}
\newcolumntype{Y}{>{\centering\arraybackslash}X}
\usepackage{siunitx}
\usepackage{vcell}

\usepackage{tikz}
\usepackage{wrapfig}
\usepackage{float}
\usepackage{caption}

\usepackage{tcolorbox}
\usepackage{listings}
\usepackage{courier}

\usepackage{xspace}

\newcommand{\method}{\texttt{\textbf{E2HiL}}\xspace}

\usepackage{amsthm}

\usepackage{algorithm}
\usepackage[noend]{algpseudocode}

\usepackage{enumitem}

\usepackage{url}
\usepackage{soul}
\usepackage{upgreek}
\usepackage{mwe}
\usepackage{calc}
\usepackage{pifont}
\usepackage{marvosym}
\usepackage{cite}

\usepackage{hyperref}

\definecolor{DarkGreen}{RGB}{0,128,0}
\definecolor{mymagenta}{HTML}{E1208A}

\title{E2HiL: Entropy-Guided Sample Selection for Efficient Real-World Human-in-the-Loop Reinforcement Learning}
\author{
Haoyuan~Deng$^{1}$,
Yudong~Lin$^{1}$,
Yuanjiang~Xue$^{1}$,
Haoyang~Du$^{1}$,
Qianzhun~Wang$^{1}$,
Boyang~Zhou$^{1}$, \\
Zhenyu~Wu$^{2}$,
and~Ziwei~Wang$^{1\dagger}$%
\thanks{This work was supported by the Singapore National Robotics Programme under Research Project DS-RFM M25N4N2009.}%
\thanks{$^{1}$ School of Electrical and Electronic Engineering, Nanyang Technological University, Singapore}
\thanks{$^{2}$ Beijing University of Posts and Telecommunications, Beijing, China}
\thanks{$^{\dagger}$Corresponding author: \textit{ziwei.wang@ntu.edu.sg}
}
}

\begin{document}
\maketitle

\begin{abstract}
Human-in-the-loop guidance has emerged as an effective approach for accelerating online reinforcement learning (RL) in real-world manipulation. 
However, existing human-in-the-loop RL (HiL-RL) frameworks often suffer from low sample efficiency, requiring substantial human interventions to achieve convergence and thereby leading to high labor costs.
To address this, we propose a sample-efficient real-world human-in-the-loop RL framework named \method, which requires fewer human interventions by actively selecting informative samples.
Specifically, stable reduction of policy entropy enables improved trade-off between exploration and exploitation with higher sample efficiency. We first build influence functions of different samples on the policy entropy, which is efficiently estimated by the covariance of action probabilities and soft advantages of policies. Then we select samples with moderate values of influence functions, where shortcut samples that induce sharp entropy drops and noisy samples with negligible effect are pruned.
Extensive experiments across 10 real-world manipulation tasks, spanning multiple embodiments and learning frameworks, demonstrate that \method improves success rates by 24.9\% while reducing human interventions by 9.3\% compared to state-of-the-art HiL-RL baselines. 
These results validate its effectiveness as a policy- and embodiment-agnostic plug-and-play module for efficient real-world RL.  
The project page can be found at \url{https://e2hil.github.io/}.
\end{abstract}
\begin{IEEEkeywords}
Human-in-the-loop reinforcement learning, robotic manipulation, sample efficiency.
\end{IEEEkeywords}

\section{Introduction}

\IEEEPARstart{R}{obotic} manipulation is a fundamental capability for robots deployed in both industrial assembly and household services~\cite{zhang2024empowering}. 
Despite decades of progress, achieving human-level manipulation skills in dynamic and unstructured environments remains a challenge~\cite{luo2024precise}. 
Meanwhile, conventional manually designed controllers struggle to generalize across diverse tasks and high-dimensional policy spaces,
motivating the transformation toward learning methods.

Although pre-trained imitation policies show strong performance, they still struggle with out-of-distribution (OOD) states~\cite{deng2025survey, lu2025vla, hung2025nora,guo2025vla}.
To improve robustness, recent work leverages RL’s trial-and-error exploration to acquire more reliable task-specific skills.
\begin{figure}[htbp]
  \centering
  \includegraphics[width=0.49\textwidth]{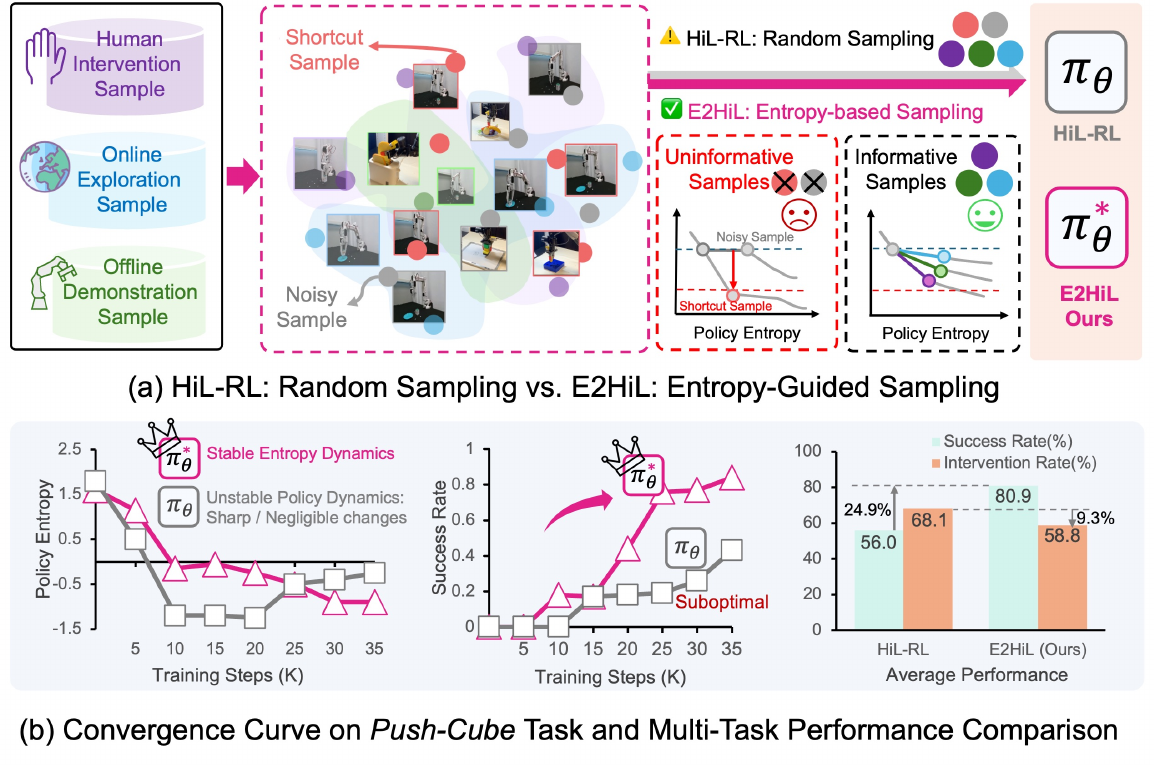}
  \caption{Compared with random uniform sampling, \method avoids early entropy collapse through entropy-guided sample selection, achieving a better exploration–performance trade-off and improving success rates with fewer human costs.}
  \label{fig:motivation}
  \vspace{-0.5cm}
\end{figure}
However, RL trained purely in simulation often fails to transfer to real robots due to gaps in dynamics and perception~\cite{yang2023sim}, prompting the adoption of real-world reinforcement learning.
Nevertheless, directly training on real-world robots intensifies challenges such as low sample efficiency~\cite{chen2024rlingua} and sparse rewards~\cite{zhang2021reinforcement, torne2024reconciling}.
Human-in-the-loop RL (HiL-RL)~\cite{luo2024precise} leverages human corrective takeovers to accelerate online policy learning.
However, existing human-in-the-loop RL methods still require extensive human intervention samples to converge, as the complexity of policy learning grows exponentially with the state–action dimensionality and task horizon, ultimately leading to high human labor costs~\cite{muraleedharan2025selective}.
This naturally raises the question: \emph{How can we achieve more stable and sample-efficient human-in-the-loop RL in real-world robotic manipulation?}

In this paper, we propose an efficient real-world human-in-the-loop RL framework named \method with active entropy-guided sample selection.
Specifically, we identify a key limitation of existing HiL-RL methods: they cannot distinguish which intervention samples are truly informative, leading to inefficient use of human effort and degraded exploration.
Fig.~\ref{fig:motivation} reveals a clear trade-off between entropy dynamics and policy improvement during HiL-RL training. This pattern mirrors recent LLM-RL findings~\cite{cui2025entropy, li2025simplevla}, where rapid entropy reduction leads to premature convergence to suboptimal policies and stable reduction of policy entropy enables improved trade-off between exploration and exploitation with higher sample efficiency.
To address this limitation, we derive influence functions that estimate effect of each sample on policy entropy using the covariance between action probabilities and soft advantages of policies.
Building on this, 
\method identifies uninformative samples with extreme values of influence functions, including shortcut samples that induce sharp entropy drops and noisy samples with negligible influence. Rather than treating all samples equally, \method\ further proposes an entropy-bounded sample selection mechanism to prune uninformative samples, which stabilizes policy entropy dynamics and prevents premature collapse. 
We validate \method across 10 real-world manipulation tasks spanning multiple embodiments and learning frameworks, achieving a 24.9\% higher success rate and 9.3\% fewer human interventions compared to state-of-the-art HiL-RL baselines. 
These results highlight entropy-guided sample selection as a promising and general strategy for efficient real-world HiL-RL, enabling robust skill acquisition with reduced human supervision.

Our contributions can be summarized as follows:
\begin{enumerate}
    \item We propose \method, an efficient real-world human-in-the-loop RL framework that improves sample efficiency by actively selecting informative samples, thereby reducing human intervention costs.
    \item We characterize sample-induced entropy dynamics through influence functions and apply an entropy-bounded selection rule to remove shortcut and noisy samples, thereby maintaining stable entropy reduction for efficient learning.
    \item We evaluate \method across 10 real-world tasks and show that its sampling efficiency enables faster convergence under reduced supervision while achieving an effective trade-off between exploration and performance.
\end{enumerate}
\section{Related Work}
\label{sec:citations}

\subsection{Real-world RL for Robotic Manipulation}
Real-world reinforcement learning (RL) aims to train manipulation policies directly on physical robots, enabling them to acquire skills that operate reliably under real sensor feedback and physical dynamics.
However, it remains highly challenging due to issues of sample efficiency~\cite{zhang2021reinforcement, ju2022transferring, torne2024reconciling}, sparse rewards~\cite{ju2022transferring, tang2025deep}, and the high cost of human interventions~\cite{luo2024precise, spencer2022expert}. To address sample inefficiency, prior works have explored off-policy and model-based RL~\cite{clavera2018model, lee2020guided}, hybrid methods~\cite{zhang2021reinforcement,pinosky2023hybrid,lei2025rl}, and offline pretraining with foundation-model priors~\cite{zhong2024empowering, kelly2019hg, kawaharazuka2024real}. Approaches such as reward shaping, inferring rewards from demonstrations or success classifiers~\cite{memarian2021self}, and leveraging video-based signals~\cite{cao2024survey} have been proposed to solve the challenge of sparse rewards. To reduce human intervention costs, reset-free learning~\cite{gupta2021reset, yang2024reset} and autonomous exploration strategies~\cite{dromnelle2023reducing} have been introduced.
However, these approaches often significantly increase training time and therefore have not achieved widespread adoption in real-world robotic learning.
Despite these advances, state-of-the-art methods increasingly adopt HiL-RL, where systems such as HIL-SERL~\cite{luo2024precise} leverage human corrective samples to rapidly acquire precise and dexterous manipulation skills.
However, conventional HiL-RL methods treat all intervention samples uniformly, making them unable to distinguish informative guidance from noisy or unhelpful corrections. This leads to inefficient use of human effort and slow policy improvement. 
In contrast, \method addresses this issue by actively selecting the informative samples to guide learning more effectively.

\subsection{Entropy-Regularized Reinforcement Learning}
Recent robotic reinforcement learning widely adopts entropy-regularized objectives, most notably through SAC~\cite{pmlr-v80-haarnoja18b}, MPO~\cite{abdolmaleki2018maximum} and RLPD~\cite{ball2023efficientonlinereinforcementlearning}. These methods show that entropy bonuses improve exploration and stabilize policy learning in manipulation tasks, but they rely on implicit temperature tuning and do not model entropy at the sample level.
Token-level entropy regularization has recently been widely adopted in the training of large language models (LLM) with RL, where regulating entropy dynamics is crucial for balancing exploration and exploitation~\cite{cheng2025reasoning, tiapkin2023fast}. Recent studies have proposed methods such as Clip-Higher~\cite{li2025simplevla} and staged temperature scheduling~\cite{renze2024effect} to encourage exploration, as well as covariance-based regularizers like Clip-Cov and KL-Cov~\cite{cui2025entropy} to mitigate entropy collapse and stabilize learning. In contrast, fine-grained entropy regularization remains largely unexplored in robotic RL, and the fundamental role of entropy in policy learning is still unclear~\cite{ball2023efficientonlinereinforcementlearning}. By analyzing action-level entropy dynamics, we observe patterns similar to those reported in LLM-RL, but with domain-specific differences~\cite{eysenbach2021maximum}. To our knowledge, \method is the first to introduce sample-level entropy regularization into real-world robotic RL, enabling more sample-efficient learning with faster convergence and reduced human intervention.


\section{Methodology}
\label{sec:method}

\subsection{Preliminaries and Problem Statement}
\label{sec:preliminaries}
We formulate real-world robotic manipulation RL as a Markov Decision Process~(MDP) $\mathcal{M} = \{\mathcal{S}, \mathcal{A}, \rho, \mathcal{P}, r, \gamma\}$, where $\mathcal{S}$ denotes the state space~(e.g., visual and proprioceptive observations), $\mathcal{A}$ is the action space~(e.g., end-effector motions), and $\rho(s_0)$ is the initial state distribution. The transition function $\mathcal{P}: \mathcal{S} \times \mathcal{A} \to \Delta(\mathcal{S})$ captures the stochastic dynamics of the system, while the reward function $r : \mathcal{S} \times \mathcal{A} \rightarrow \mathbb{R}$ specifies task objectives. 
The discount factor $\gamma \in [0,1)$ balances immediate and long-term return. We denote by $s_t \in \mathcal{S}$ and $a_t \in \mathcal{A}$ the state and action at time $t$.

\begin{figure*}[htbp]
  \centering
  \includegraphics[width=1.0\textwidth]{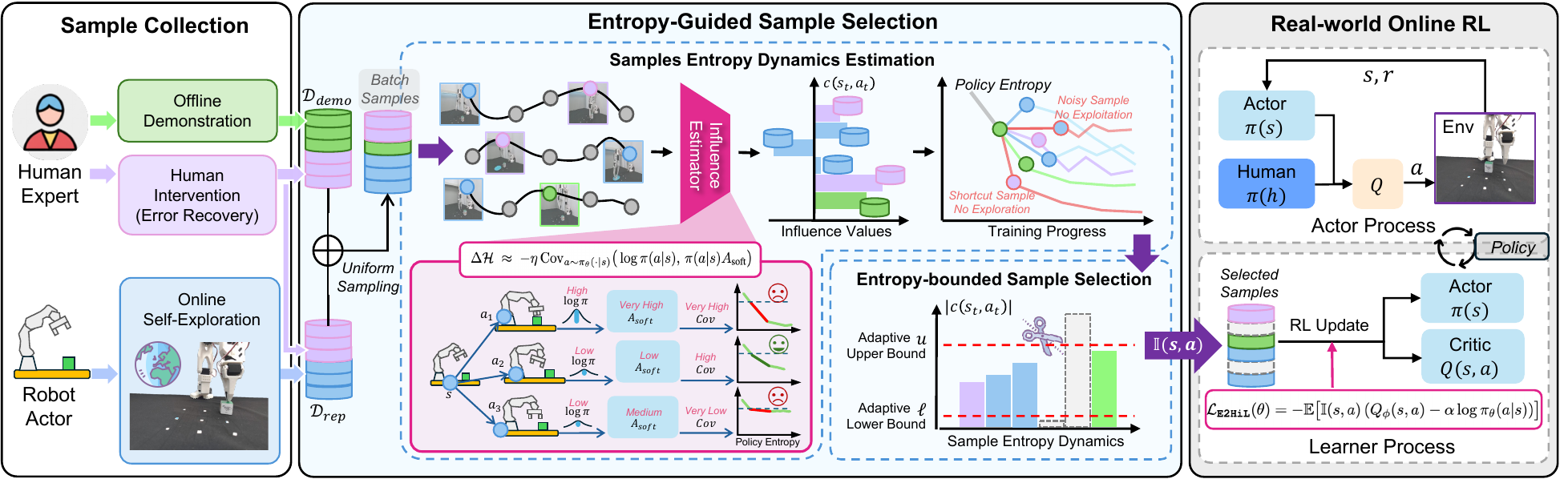}
  \caption{\textbf{Entropy-Guided Sample Selection Framework.}
Our framework consists of two key components: Sample Entropy Dynamics Estimation, where we characterize the entropy dynamics induced by each sample, and Entropy-Bounded Sample Selection, where we prune shortcut and noisy samples by constraining their influence value within dynamic bounds.}
\vspace{-0.3cm}
  \label{fig:pipeline}
\end{figure*}


To enhance real-world training efficiency, we employ human-in-the-loop reinforcement learning to leverage both human intervention and autonomous exploration samples for improved sample efficiency~\cite{ball2023efficientonlinereinforcementlearning}:
\begin{equation}
\label{1}
\mathcal{L}_Q(\phi) = 
\mathbb{E}_{(s_t, a_t, r_t, s_{t+1})\sim \mathcal{D}}
\Big[ \big(Q_\phi(s_t, a_t) - \hat{y}_t \big)^2 \Big]
\end{equation}
where the regression target $\hat{y}_t$ 
is the bootstrapped Bellman target estimating the expected return under the current policy. 
A slowly updated target network is used to compute $\hat{y}_t$ to stabilize training.
In practice, policy is parameterized by neural networks $\pi_\theta(a_t \mid s_t)$ and the actor parameters $\theta$ are updated to maximize the entropy-regularized objective:
\begin{equation}
\label{equation3}
\mathcal{L}_\pi(\theta) = 
-\mathbb{E}_{(s_t, a_t) \sim \mathcal{D}}
\Big[ Q_\phi(s_t, a_t) - \alpha \log \pi_\theta(a_t \mid s_t) \Big]
\end{equation}
where $\alpha$ is an adaptive temperature coefficient controlling the balance between exploitation and exploration. 

However, conventional HiL-RL methods~\cite{luo2024precise} uniformly sample from replay buffers with high human intervention cost. According to empirical results in \cite{cui2025entropy}, stable reduction entropy can improve the exploration and exploitation trade-off with higher sample efficiency, which can reduces the human intervention cost significantly. Therefore, we explicitly formulate our learning objective to incorporate \emph{entropy-guided active sample selection}, wherein training samples contributing to moderate entropy reduction are selected. 
The objective for HiL-RL is written as follows:
\begin{equation}
\label{eq:ours_objective}
\mathcal{L}_{\text{\method}}(\theta) =
-\mathbb{E}
\big[ \mathbb{I}(s_t,a_t)\, (Q_\phi(s_t,a_t) - \alpha \log \pi_\theta(a_t \mid s_t)) \big]
\end{equation}
where $\mathbb{I}(s_t,a_t) \in \{0,1\}$ is an indicator function 
determined by the entropy-bounded selection criterion. 
Samples with excessive or negligible entropy contribution are masked out ($\mathbb{I}=0$), removing their gradients from the update. 
This selective objective preserves only entropy-consistent samples, thereby reducing variance and improving sample efficiency.


\subsection{Overall Pipeline}
Our proposed framework aims to stabilize policy entropy dynamics in human-in-the-loop RLPD learning by explicitly regulating the selection and use of replay samples during training. 
As illustrated in Fig.~\ref{fig:pipeline}, the entire pipeline follows an iterative loop of sample collection, entropy-guided active sample selection, and online RL policy update.
For sample collection, \method uses three sources: offline demonstrations before training, robot self-exploration samples during online interaction, and human intervention samples which are corrective takeover for error recovery. All collected transitions \((s_t, a_t, r_t, s_{t+1})\) are stored in the replay buffer \(\mathcal{D}_{rep}\) and demonstration buffer \(\mathcal{D}_{demo}\).
Before updating the policy, we estimate influence functions of different samples on the policy
entropy, which is formulated by influence value \(c(s_t,a_t)\): the covariance of
action probabilities and soft advantages of policy.
Then, we propose entropy-bounded sample selection mechanism that
assigns a binary indicator \(\mathbb{I}(s_t,a_t)\) to retain samples with moderate influence.
Samples with moderate influence on entropy are retained for policy updates, while those causing sharp drops or negligible changes are excluded. This entropy-guided selection stabilizes entropy dynamics and helps maintain a proper balance between exploration and exploitation. By integrating entropy estimation with selective sampling in a closed training loop, the framework improves sample efficiency and enhances convergence stability in real-world human-in-the-loop RL.


\subsection{Sample Entropy Dynamics Estimation}

The key challenge in human-in-the-loop RL is improving sample efficiency while avoiding unstable entropy collapse caused by uniform replay. 
We therefore quantify each sample’s effect on policy entropy via a batch-wise influence function derived from entropy dynamics. 
Specifically, the entropy change of policy $\pi_\theta$ can be approximated as~\cite{cui2025entropy}:
\begin{equation}
\label{dynamics}
\Delta \mathcal{H} \approx
\mathbb{E}_{s_t \sim d_{\pi_\theta}}
\Big[-\,\mathrm{Cov}_{a_t \sim \pi_\theta}
\big(\log \pi_\theta(a_t \mid s_t),\, \Delta z_t\big)\Big],
\end{equation}
where $\Delta z_t$ denotes the logit update for action $a_t$ at state $s_t$, and $\mathrm{Cov}(\cdot,\cdot)$ is taken under $\pi_\theta(\cdot \mid s_t)$. 
A detailed derivation of this entropy-dynamics approximation is provided in the Appendix~II.
Intuitively, by relating policy updates to changes in entropy, Eq.~\eqref{dynamics} provides a measurable way to assess each sample's influence on policy entropy.

\textbf{Logit Update under RLPD:}
To make Eq.~\eqref{dynamics} applicable in practice, a straightforward approach is to express $\Delta z_{t}$ using the actor gradient in RLPD, where policy logits are updated through the standard policy gradient:
\begin{equation}
\label{equation5}
\Delta z_{t} \;\coloneqq\; z^{t+1}_{t} - z^{t}_{t}
\;=\; -\,\eta \, \nabla_{\theta} \mathcal{L}_\pi(\theta)
\end{equation}
where the negative sign arises from minimizing the RLPD actor objective function $\mathcal{L}_\pi$ via gradient descent, and  $\eta$ denotes the learning rate.  
Let \(g(s_t,a_t') \triangleq Q_\phi(s_t,a_t') - \alpha \log \pi_\theta(a_t' \mid s_t)\).
Here \(a_t\) denotes the specific action index associated with the logit parameter \(\theta_{s_t,a_t}\), 
while \(a_t'\) denotes actions sampled from the policy \(\pi_\theta(\cdot \mid s_t)\) inside the expectation.
Using the score-function, the gradient of the actor objective is:
\begin{align}
\nabla_{\theta} \mathcal{L}_\pi(\theta)
&= -\,\pi_\theta(a_t \mid s_t)\,
\Big(\underbrace{g(s_t,a_t) - V_\pi(s_t)}_{A_{\text{soft}}(s_t,a_t)}\Big)
\label{equation10}
\end{align}
See Appendix~I for full derivation.
Here \(A_{\text{soft}}(s_t,a_t)\) provides an approximation of the off-policy soft advantage, and 
\(V_\pi(s_t)\) is defined as the entropy-regularized state value:
\begin{equation}
V_\pi(s_t)
\;\coloneqq\;
\mathbb{E}_{a_t' \sim \pi_\theta}
\big[ Q_\phi(s_t,a_t') - \alpha \log \pi_\theta(a_t' \mid s_t) \big]
\label{equation_value}
\end{equation}
Eq.~\eqref{equation10} naturally appears in RLPD when expressing the policy gradient in terms of the soft advantage, making the actor update depend on the action's value relative to the entropy-regularized state value.
Then we obtain the policy logit change for action \(a_t\) at state \(s_t\):
\begin{equation}
\label{eq:logit-step}
\Delta z_{t}
= -\,\eta \,\nabla_{\theta} \mathcal{L}_\pi(\theta)
= \eta\,\pi_\theta(a_t \mid s_t)\,A_{\text{soft}}(s_t,a_t)
\end{equation}
This decomposition expresses logit updates in terms of policy probabilities and soft advantages, enabling a sample-wise characterization of their influence on entropy.

\textbf{Policy Entropy Change Estimation:}
Substituting Eq.~\eqref{eq:logit-step} into Eq.~\eqref{dynamics}, the influence function approximation yields:
\begin{equation}
\label{eq:entropy-cov}
\Delta\mathcal{H}
\;\approx\;
-\,\eta\,
\mathrm{Cov}
\big(\log \pi_\theta(a_t \mid s_t),\;
\pi_\theta(a_t \mid s_t)\,A_{\text{soft}}(s_t,a_t)\big).
\end{equation}
which indicates that entropy dynamics are determined by the covariance between log-probabilities and the soft advantage of policies. This reflects how different samples influence policy entropy: actions with high probability and large soft advantage tend to reduce entropy, while actions with low probability and large soft advantage tend to increase it.

\begin{algorithm}[t]
  \caption{RL with Entropy-guided Sample Selection}
  \label{algo:simple}
  \begin{algorithmic}
    \State Temperature $\alpha$, gradient steps $G$, batch size $N$
    \State Initialize critics $\phi_i$ (targets $\phi_i' \gets \phi_i$); initialize actor $\theta$
    \State Initialize replay buffer $\mathcal D_{\text{rep}}$ and demo buffer $\mathcal D_{\text{demo}}$
      \State Receive initial state $s_0$
      \For{$t=0$ \textbf{to} $T$}
        \If{human intervention}
          \State Take action $a_t$ from human operator
          \State Store $(s_t,a_t,r_t,s_{t+1})$ in $\mathcal D_{\text{rep}}$ and $\mathcal D_{\text{demo}}$
        \Else
          \State Sample $a_t \sim \pi_\theta(\cdot\mid s_t)$
          \State Store $(s_t,a_t,r_t,s_{t+1})$ in $\mathcal D_{\text{rep}}$
        \EndIf
        \For{$g=1$ \textbf{to} $G$}
          \State Sample minibatch $b_R$ of $N/2$ from $\mathcal D_{\text{rep}}$
          \State Sample minibatch $b_D$ of $N/2$ from $\mathcal D_{\text{demo}}$
          \State Set $b \gets b_R \cup b_D$
          \State Update $\phi$ minimizing loss from Eq.~\eqref{1}:
          \State Update target networks $\phi_i' \leftarrow \rho\phi_i' + (1-\rho)\phi_i$
        \EndFor
        \State \textcolor{mymagenta}{Estimate influence value $c_i$ on $b$ from Eq.~\eqref{equation18}}
        \State \textcolor{mymagenta}{Get indicator function $\mathbb{I}(s_t,a_t)$ from Eq.~\eqref{equation19}}
        \State \textcolor{mymagenta}{Update actor by maximizing objective Eq.~\eqref{eq:actor-clipcov}}
      \EndFor
  \end{algorithmic}
\end{algorithm}
\vspace{-0.4cm}

\subsection{Entropy-guided Sample Selection}

To stabilize policy entropy and better utilize costly human interventions, we propose an \textit{entropy-guided sample selection} framework, with two components: \textit{sample entropy dynamics estimation} and \textit{entropy-bounded sample selection}.

\textbf{Sample Entropy Dynamics Estimation:} 
Given a state--action pair \((s_t, a_t)\), we estimate the covariance term in 
Eq.~\eqref{eq:entropy-cov} via Monte Carlo sampling over 
\(\pi_\theta(\cdot \mid s_t)\).
Specifically, for each state \(s_t\), we draw \(K\) actions 
\(\{a_{t,k}\}_{k=1}^K \sim \pi_\theta(\cdot \mid s_t)\) and compute the covariance.
We define the influence value of sample \((s_t,a_t)\) on entropy dynamics as
\begin{equation}
c(s_t, a_t)
\;\triangleq\;
-\,\eta\,\widehat{\mathrm{Cov}}(s_t, a_t).
\label{equation18}
\end{equation}
The quantity \(c(s_t,a_t)\) is treated as a stop-gradient signal that reflects how sample 
contributes to the entropy changes.

\textbf{Entropy-bounded Sample Selection:} 
In practice, we observe that a small portion of samples exhibit extremely large or small covariance magnitudes, far beyond the range of the majority within the same batch.  
These outlier samples play a dominant role in triggering abnormal entropy fluctuations, which can either drive the policy into premature collapse or stall its progress.  
Concretely, we categorize some problematic samples: (1) shortcut samples that induce abrupt entropy drops and cause premature policy collapse, and (2) noisy samples that contribute negligibly to entropy change and slow down learning progress~(see Sec.~\ref{exp:dynamics} for empirical verification). 
To mitigate the impact of samples that cause abrupt entropy changes or contribute negligibly to exploration, 
we clip the estimated absolute influence value $\{|c(s_t,a_t)|\}$ within a dynamically adjusted interval 
$[\ell, u]$,
where the bounds are computed adaptively from the batchwise percentile range of influence magnitudes.
Formally, we define an indicator function that determines whether a sample participates in the policy update:
\begin{equation}
\mathbb{I}(s_t,a_t) = \mathbf{1}\big[\,|c(s_t,a_t)| \in [\ell, u]\,\big].
\label{equation19}
\end{equation}
Samples outside this range are masked out ($\mathbb{I}=0$), effectively canceling their gradient contributions.
Given the per-sample actor loss $
l_t \;\coloneqq\; Q_\phi(s_t,a_t)\;-\;\alpha \log \pi_\theta(a_t \mid s_t)$, our entropy-aware actor objective is defined as:
\begin{equation}
\label{eq:actor-clipcov}
\mathcal{L}_{\mathrm{\method}}(\theta)
\;=\;
-\,
\frac{
\sum_{(s_t,a_t)\in \mathcal{B}}
\mathbb{I}(s_t,a_t)l_t
}{
\sum_{(s_t,a_t)\in \mathcal{B}}
\mathbb{I}(s_t,a_t)\;+\;\varepsilon}
\end{equation}
where $\mathcal{B}$ denotes a minibatch of size $B$, and $\varepsilon$ is a small constant for numerical stability.
This formulation ensures that only entropy-consistent samples contribute to the gradient, leading to smoother entropy evolution and more stable convergence in human-in-the-loop training. 
A detailed theoretical analysis of the effects of entropy-bounded masking on bias and variance is provided in Appendix~III.

\begin{figure}[htbp]
  \centering
  \includegraphics[width=0.48\textwidth]{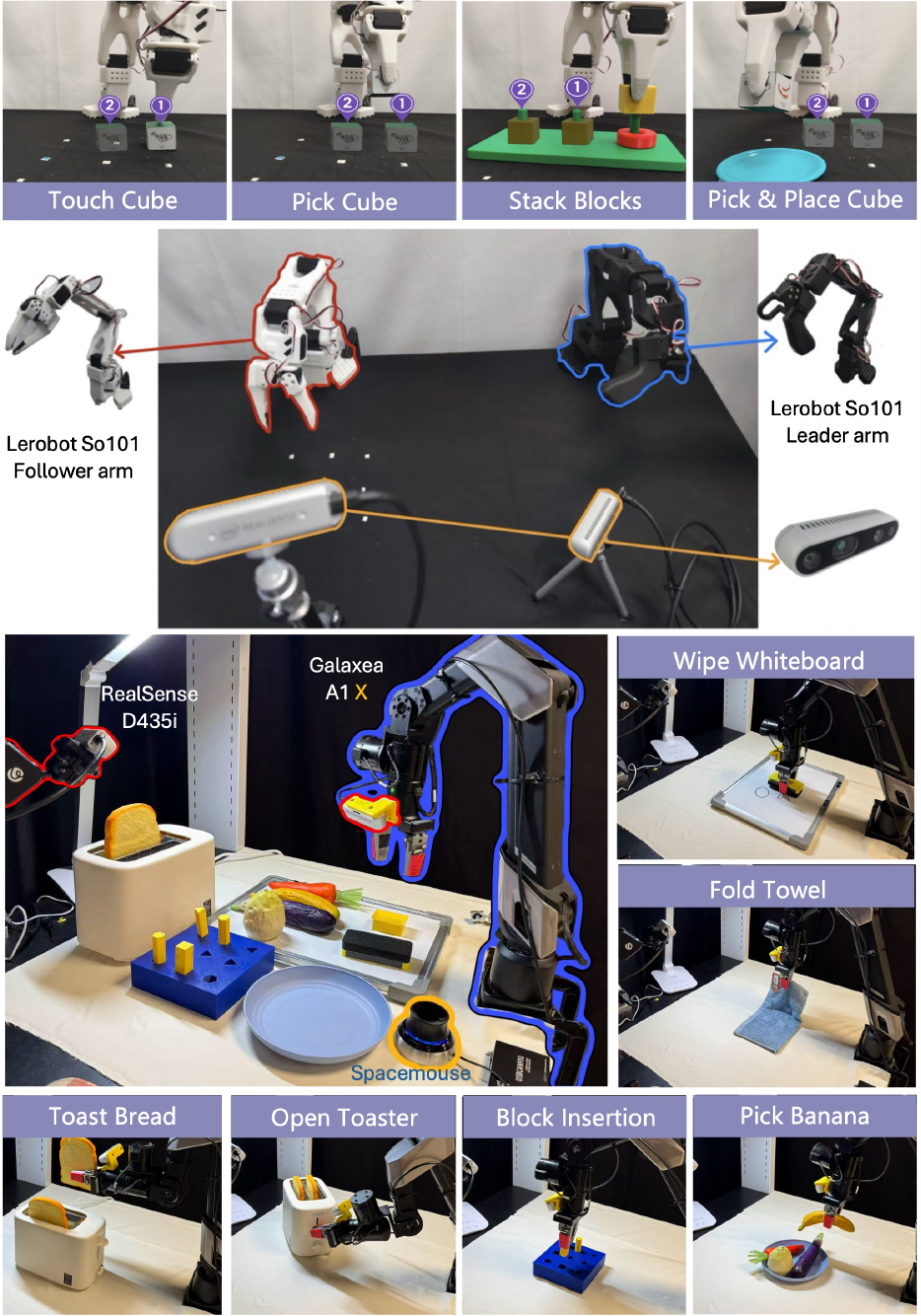}
  \caption{Real-world experimental setups on Lerobot SO-101 and A1\_X arm, comprising 10 manipulation tasks.}
  \vspace{-0.6cm}
  \label{fig:robot setup}
\end{figure}

\section{Experiments}
\label{sec:result}

In this section, we first describe the real-world experimental setups and compare \method with state-of-the-art HiL-RL methods. 
We then validate the accuracy of our entropy-dynamics estimation and the effectiveness of entropy-guided sample selection. 
Finally, we present ablation studies and hyperparameter analyses to examine the influence of key design choices, including sample pruning behavior, intervention types, and entropy-related parameters, on policy learning in real-world robotic manipulation.

\subsection{Experimental Setup}
\textbf{Baseline.} 
We adopt HIL-SERL~\cite{luo2024precise} and ConRFT~\cite{chen2025conrft} as our primary baselines for real-world human-in-the-loop manipulation. 
HIL-SERL trains policies from scratch with corrective human interventions and entropy regularization from RLPD to stabilize updates. 
ConRFT initializes policies using Cal-QL~\cite{nakamoto2024calqlcalibratedofflinerl} and behavior cloning pretraining, followed by RL fine-tuning of vision-language-action policies in a multi-stage paradigm. 
Both methods are evaluated under identical cross-embodiment settings on multi-object, long-horizon, contact-rich, and non-rigid tasks (Fig.~\ref{fig:robot setup}). Detailed task descriptions are provided in the Appendix~IV.

\textbf{Evaluation Metrics.} 
We report three main metrics: the final \textbf{success rate} and the average \textbf{human intervention rate}, which is defined as the ratio between the number of intervention steps and the total robot interaction steps.
We also monitor the evolution of the \textbf{policy entropy curve} to evaluate the stability of learning and the quality of exploration.

\subsection{Comparisons with the State-of-the-Art}\label{subsec:comp_with_sota}
\begin{figure}[t]
  \centering
  \includegraphics[width=0.48\textwidth]{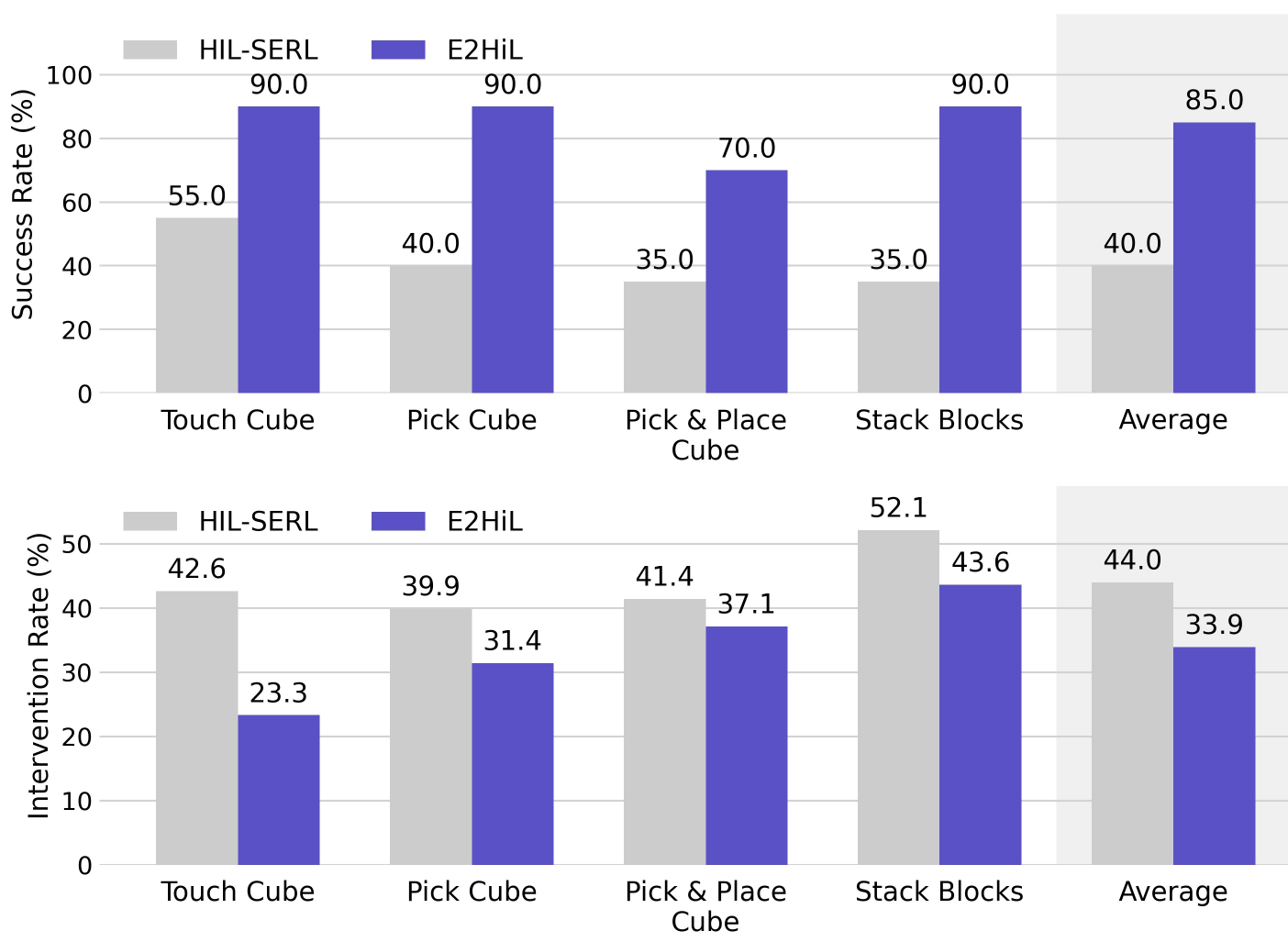}
  \caption{\textbf{Real-world Manipulation Results on the Lerobot.} \method significantly outperforms state-of-the-art human-in-the-loop RL approaches by achieving higher success rates with fewer human interventions across four tasks.}
  \vspace{-0.3cm}
  \label{fig:lerobot_main}
\end{figure}

\begin{figure}[t]
  \centering
  \includegraphics[width=0.49\textwidth]{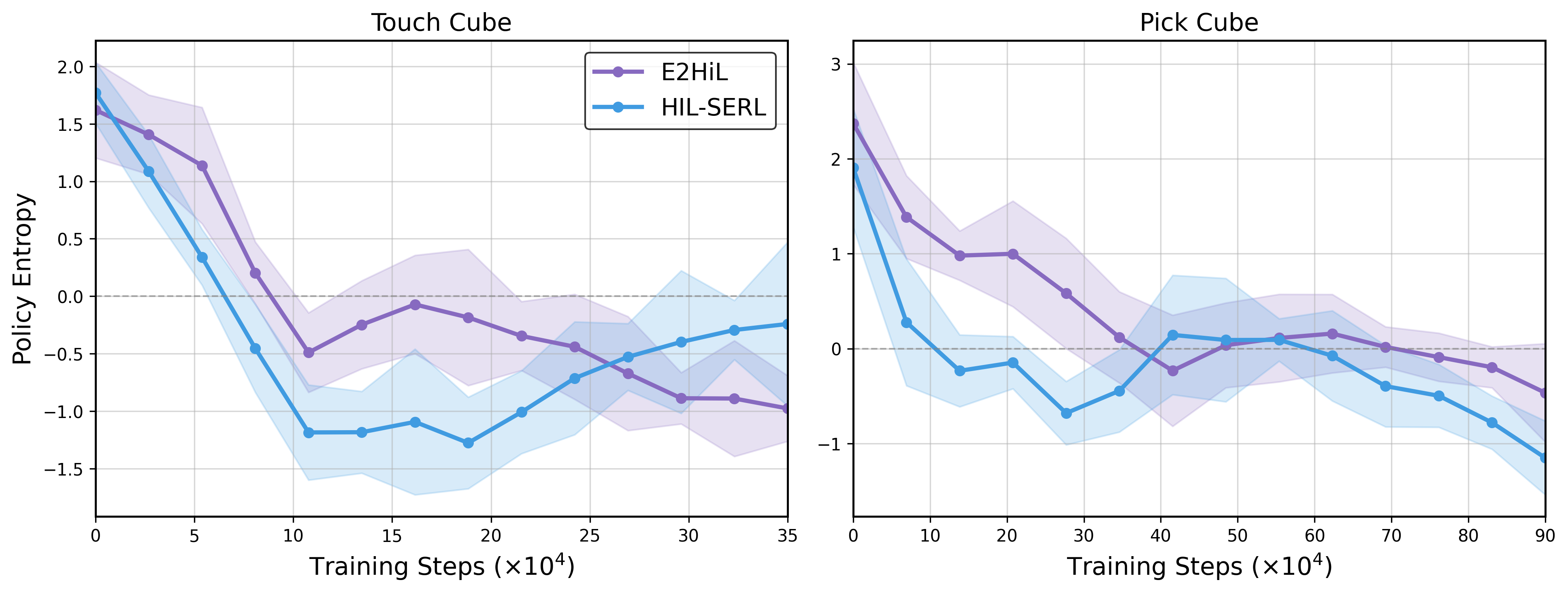}
  \caption{With entropy-guided sample selection, \method exhibits more stable entropy dynamics, avoiding premature entropy collapse and ineffective updates.}
  \vspace{-0.6cm}
  \label{fig:lerobot_line}
\end{figure}

\textbf{General Comparison on the Lerobot.} We compare \method with the state-of-the-art method HIL-SERL and report the performances across four position-generalization tasks of increasing difficulty. 
As shown in Fig.~\ref{fig:lerobot_main}, \method achieves the best performance in terms of convergence success rate, average human intervention rate, and the stability of the entropy curve. These results demonstrate that entropy-guided sample selection mechanism not only improves task performance and reduces human effort, but also stabilizes the entropy dynamics during training. From 
Fig.~\ref{fig:lerobot_line}, we observe that \method achieves a sizable relative improvement of \textbf{45\%} in success rate across four tasks, while reducing the average human intervention rate by \textbf{10.1\%}. 

\textbf{General Comparison on the A1\_X.}
We further evaluate \method on a broader set of real-world manipulation tasks using the A1\_X platform.
As shown in Table~\ref{tab:real_world_results}, \method improves the average success rate by 17.5\% while reducing human intervention by 9.4\% across all tasks. 
Table~\ref{tab:conrft} further shows consistent gains over ConRFT, with a 6.6\% increase in success rate and a 7.5\% reduction in intervention rate.
These improvements are consistently observed across different learning frameworks, indicating that \method is policy- and embodiment-agnostic, serving as a plug-and-play optimization module without architecture-specific modification.


\begin{table*}[t]

\centering
\small
\setlength{\tabcolsep}{7.5pt}
\caption{\textbf{Real-world Manipulation Results on the A1\_X robot.} 
\method significantly outperforms state-of-the-art human-in-the-loop RL approaches by achieving higher success rates with fewer human interventions across six tasks.}
\label{tab:real_world_results}

\renewcommand{\arraystretch}{1}

\begin{tabular}{llccccccc}
\toprule

\multirow{2}{*}{Method} & \multirow{2}{*}{Metrics} 
& \multicolumn{1}{c}{Multi-object}
& \multicolumn{1}{c}{Long-horizon}
& \multicolumn{3}{c}{Contact-rich}
& \multicolumn{1}{c}{Non-rigid}
& \multirow{2}{*}{Average} \\

\cmidrule(lr){3-3}
\cmidrule(lr){4-4}
\cmidrule(lr){5-7}
\cmidrule(lr){8-8}

&
& \makecell{Pick\\[-1pt]Banana}
& \makecell{Toast\\[-1pt]Bread}
& \makecell{Open\\[-1pt]Toaster}
& \makecell{Block\\[-1pt]Insertion}
& \makecell{Wipe\\[-1pt]Whiteboard}
& \makecell{Fold\\[-1pt]Towel} \\

\midrule

BC
& Success Rate \textcolor{green}{$\uparrow$}
& 20.0 & 15.0 & 30.0 & 20.0 & 30.0 & 15.0 & 21.7 \\

\midrule

\multirow{2}{*}{HiL-SERL}
& Success Rate \textcolor{green}{$\uparrow$}
& 65.0 & 55.0 & 85.0 & 60.0 & 70.0 & 65.0 & 66.7 \\
& Intervention Rate \textcolor{purple}{$\downarrow$}
& 58.6 & 68.1 & 43.2 & 54.6 & 65.3 & 66.3 & 59.4 \\

\midrule

\multirow{2}{*}{\makecell[l]{HiL-SERL\\ + \method}}
& Success Rate \textcolor{green}{$\uparrow$}
& 85.0 & 75.0 & 90.0 & 85.0 & 90.0 & 80.0 & 84.2 \\
& Intervention Rate \textcolor{purple}{$\downarrow$}
& 48.1 & 55.2 & 39.1 & 46.3 & 55.6 & 55.8 & 50.0 \\

\bottomrule
\end{tabular}
\vspace{-0.4cm}
\end{table*}
\begin{figure}[htbp]
  \centering
  \includegraphics[width=0.49\textwidth]{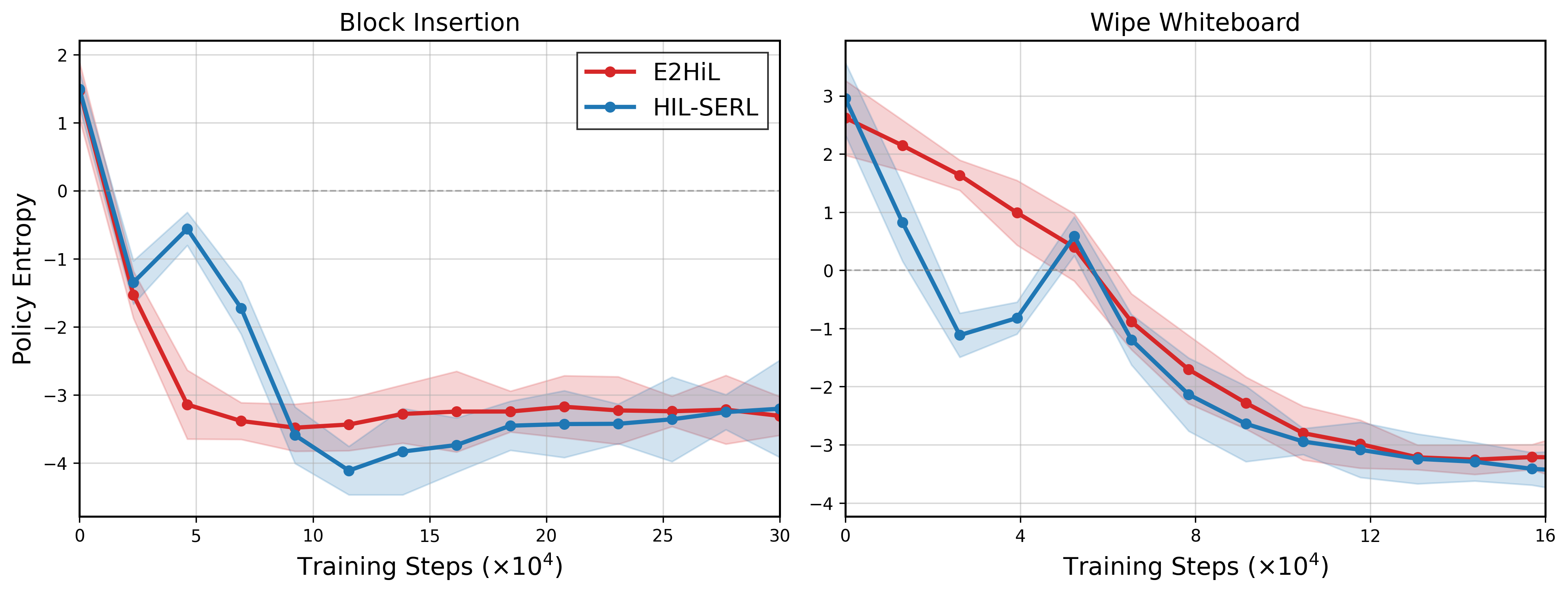}
  \caption{\method exhibits more stable entropy dynamics, avoiding premature entropy collapse}
  \vspace{-0.15cm}
  \label{fig:Hilserl_line}
\end{figure}




\begin{table}[t]
\centering
\small
\setlength{\tabcolsep}{4pt}
\renewcommand{\arraystretch}{1.1}

\caption{Comparison with ConRFT~\cite{chen2025conrft} on the A1\_X}
\label{tab:conrft}

\resizebox{\linewidth}{!}{
\begin{tabular}{lcc}
\hline
 & \multicolumn{2}{c}{Task} \\
\cline{2-3}
Metrics & Block Insertion & Pick Banana  \\
\hline
$\Delta$ Success Rate (\%) $\uparrow$ & $+$4.5 & $+$8.6 \\
$\Delta$ Intervention Rate (\%) $\downarrow$ & $-$6.7 & $-$8.3 \\
\hline
\end{tabular}
}
\vspace{-0.3cm}
\end{table}
\textbf{Case Study Analysis.}
Fig.~\ref{fig:lerobot_line} shows a case study on the \textit{Touch-Cube} task. 
During training, the cube is randomly placed at position~1 or~2 to prevent overfitting and require the agent to learn both skills, enabling evaluation of learning efficiency and generalization. 
Both \method and HIL-SERL initially reduce entropy while learning the first position. 
However, by clipping shortcut samples that would cause abrupt entropy collapse, \method maintains a smoother entropy decrease and acquires the first skill at around $10k$ steps. Because entropy remains sufficiently high, it continues exploring, discovers the second position at $20k$ steps, and acquires it by $33k$ steps. In contrast, the baseline converges more slowly and requires higher human intervention.

\subsection{Analysis of Samples Entropy Dynamics}
\label{exp:dynamics}
\begin{figure}[htbp]
  \centering
  \includegraphics[width=0.48 \textwidth]{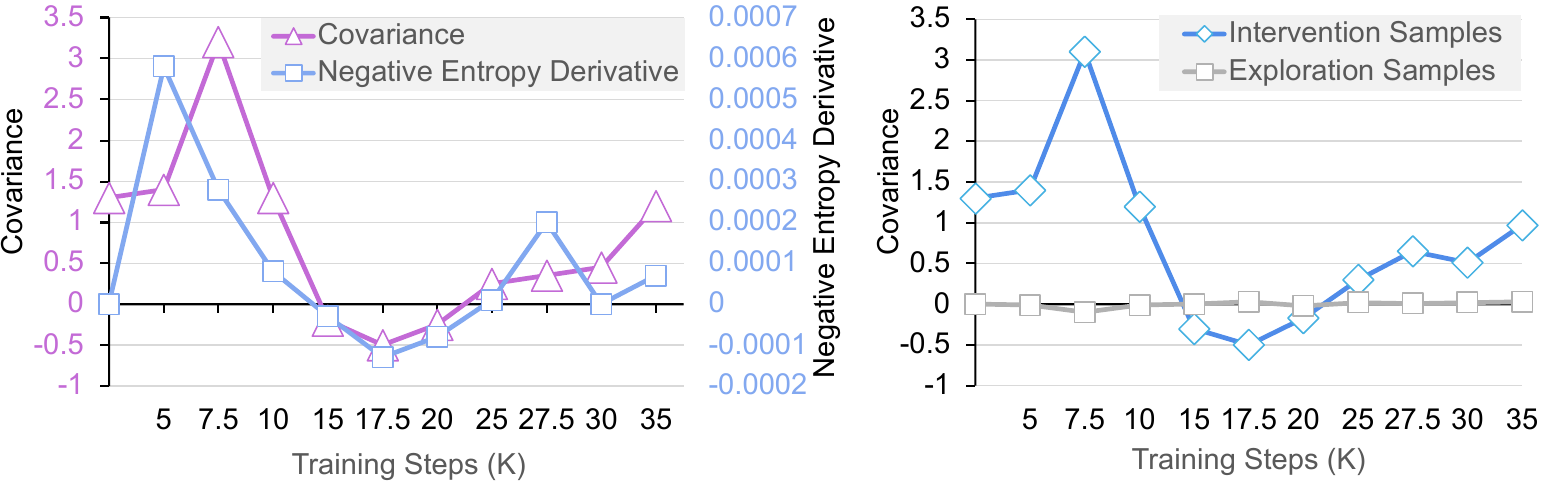}
  \caption{\textit{Left:} Policy entropy derivative and covariance during HiL-RL training on the \textit{Touch-Cube} task. \textit{Right:} Covariance of human intervention vs. self-exploration samples.}
  \label{fig:prediction}
  \vspace{-0.15cm}
\end{figure}

\begin{figure}[htbp]
  \centering
  \includegraphics[width=0.49\textwidth]{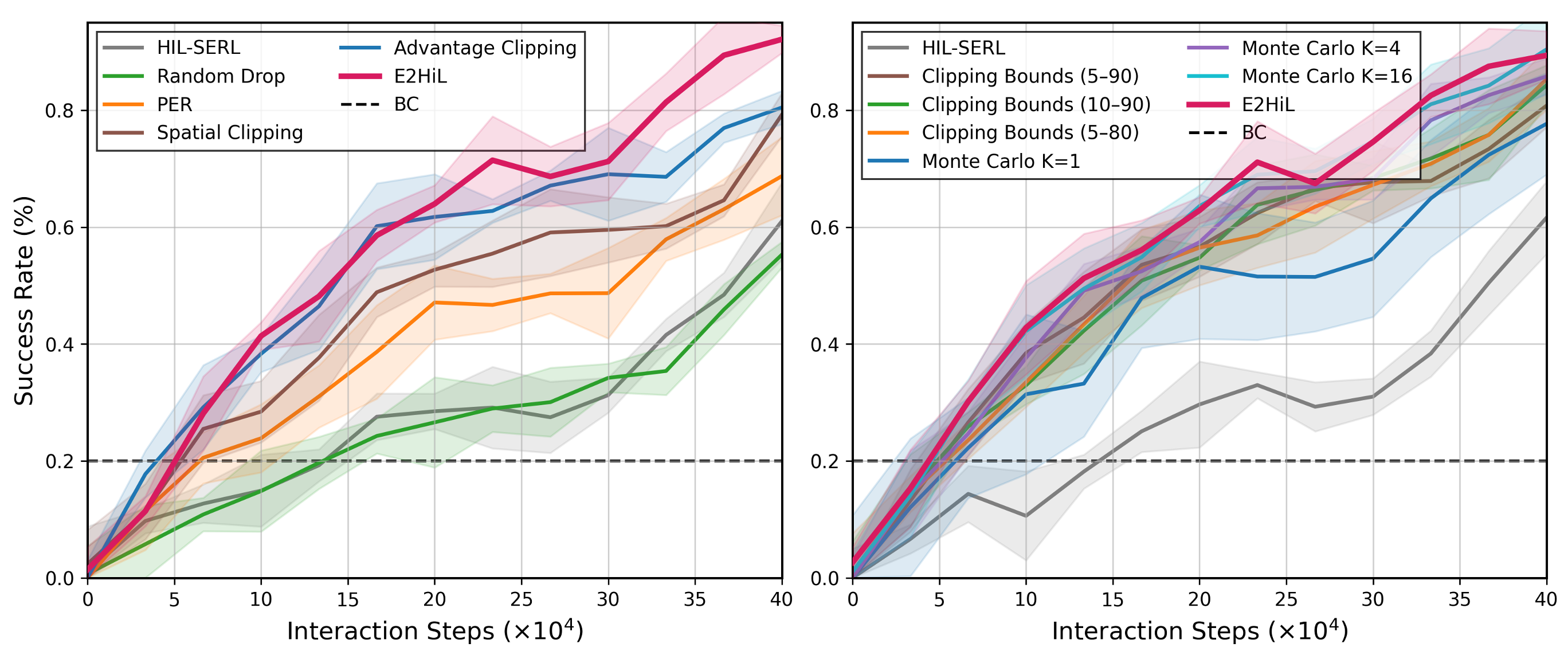}
  \caption{\textbf{Ablation and Hyperparameter Analysis.} Comparison of sampling strategies and hyperparameter settings demonstrates that \method outperforms all baselines.}
  \vspace{-0.35cm}
  \label{fig:ablation}
\end{figure}

\textbf{Empirical Verification.} 
A natural question is how accurately our method estimates the contribution of samples to policy entropy change. 
To investigate this, we empirically validate our key result that sample-induced entropy dynamics can be characterized by the covariance between action log-probabilities and soft advantages, as formulated in Eq.~\eqref{eq:entropy-cov}. 
During HIL-SERL training on the \textit{Touch-Cube} task, we compute the batch-wise average covariance $\mathrm{Cov}(\cdot)$ at each policy update and compare it with the ground-truth entropy derivative $-\Delta\mathcal{H}$. 
As shown in Fig.~\ref{fig:prediction}~\textit{Left}, the two curves exhibit a strong proportional relationship 
consistent with our theory, particularly during the initial entropy reduction phase (0--$10k$ steps) and the subsequent entropy increase ($10k$--$20k$ steps). 
After $20k$ steps, when policy entropy stabilizes, covariance values fluctuate around zero, further supporting the validity of our estimation. 
Deviations observed in the first $5k$ steps are likely caused by inaccurate critic Q-value estimates, which bias the computation of $A_{\text{soft}}$ in Eq.~\eqref{eq:entropy-cov}.
\begin{table}[t]
\centering
\caption{\textbf{Samples Covariance Distribution.} Mean absolute covariance across different sample types.}
\label{tab:abs_cov_means}
\resizebox{1\linewidth}{!}{
\begin{tabular}{c|c|c|c}
\toprule
\textbf{Subset} & \textbf{All Samples} & \textbf{Intervention Samples} & \textbf{Exploration Samples} \\
\midrule
\rowcolor{blue!5}
Top 2\% & 269.9 & \color{blue!50}368.1 & 5.56 \\
Top 5\% & 127.0 & \color{blue!50}180.2 & 2.7 \\
\rowcolor{blue!5}
Top 10\% & 67.3 & \color{blue!50}98.6 & 1.5 \\
Top 20\% & 34.2 & \color{blue!50}51.1 & 0.8 \\
\rowcolor{blue!5}
Top 50\% & 13.8 & \color{blue!50}20.6 & 0.3 \\
Low 20\% & 0.001 & \color{blue!50}0.002 & 0.0007 \\
\rowcolor{blue!5}
Low 10\% & 0.0005 & \color{blue!50}0.0007 & 0.0003 \\
Low 5\% &0.0002 & \color{blue!50}0.0003 & 0.0001 \\
\rowcolor{blue!5}
Low 2\% & 0.00008 & \color{blue!50}0.0001 & 0.00006 \\
All & 6.9 & \color{blue!50}10.3 & 0.2 \\
\bottomrule
\end{tabular}}
\vspace{-0.5cm}
\end{table} As shown in Fig.~\ref{fig:ablation}, although early-stage Q inaccuracies may affect covariance estimation, they have negligible impact on final task success, indicating that \method can be applied in a zero-shot manner without additional calibration.

\textbf{Human Intervention Sample Analysis.}
Having verified the accuracy of our entropy-dynamics characterization, we next analyze how human intervention and self-exploration samples differ in their influence on policy learning in real-world HiL-RL. 
During HIL-SERL training on the \textit{Touch-Cube} task, we compute the per-sample covariance $c_i$ and tag samples originating from human interventions. 
As shown in Fig.~\ref{fig:prediction}~\textit{Right}, human intervention samples exhibit significantly larger amplitudes and higher mean covariance values than self-exploration samples, indicating stronger influence on entropy dynamics. 
This observation is supported by Table~\ref{tab:abs_cov_means}, which reports covariance magnitude distributions over all samples and the two subsets. Human intervention samples dominate the upper tail of the distribution, suggesting a decisive role in driving entropy dynamics and accelerating early policy improvement. However, both sample types also introduce instability. We identify two problematic cases: 
\begin{figure*}
   \centering
   \includegraphics[width=1\textwidth]{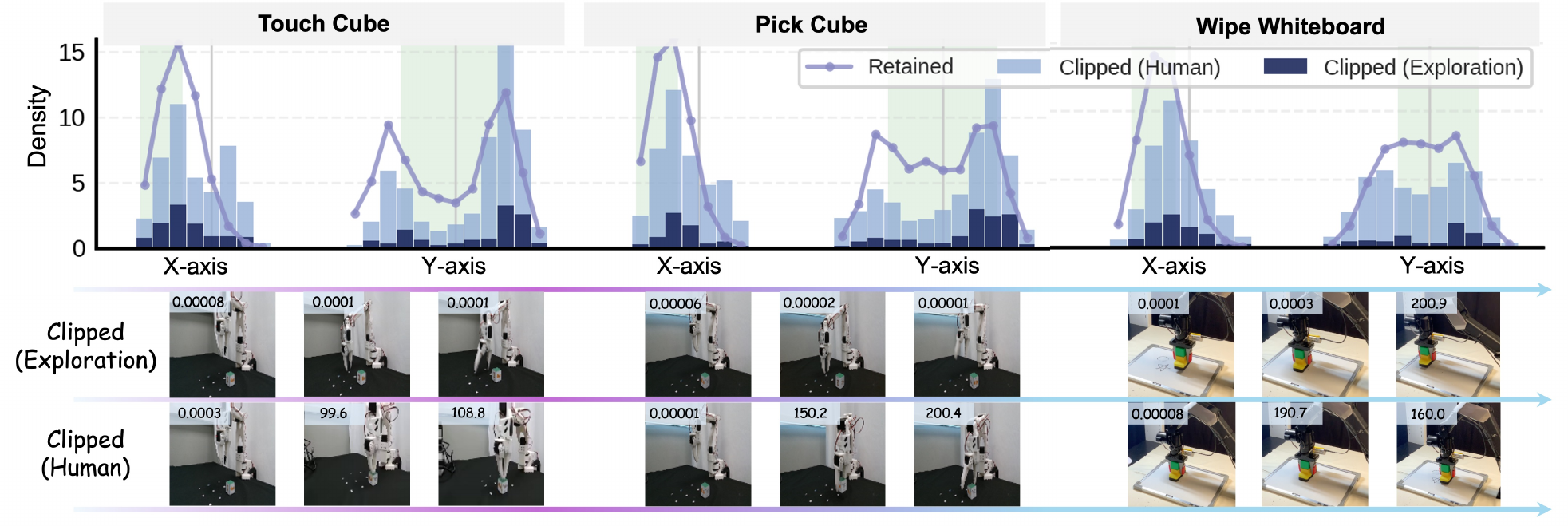}
  \caption{\textbf{Retained vs.~Clipped Sample Distribution in Three Tasks.} Line density of retained samples and histogram of clipped samples shown in task space (gripper position). Most clipped samples come from human interventions, and their covariance visualization highlights out-of-workspace or redundant states.}
  \label{fig:touch_histogram}
  \vspace{-0.4cm}
\end{figure*}
\begin{table}[t]
\centering
\small
\setlength{\tabcolsep}{4pt}
\renewcommand{\arraystretch}{1.1}

\caption{Comparison with Q-estimation Variants.}
\label{tab:Q_variant_ablation}

\resizebox{\linewidth}{!}{
\begin{tabular}{lccc}
\hline
 & \multicolumn{3}{c}{Variant} \\
\cline{2-4}
Metrics & E2HiL & E2HiL-Cal-QL & E2HiL-Ensemble \\
\hline
Success Rate (\%) $\uparrow$ & \underline{85} & \textbf{90} & \underline{85} \\
Intervention Rate (\%) $\downarrow$ & \textbf{46.3} & 48.1 & \underline{47.8} \\
\hline
\end{tabular}
}
\vspace{-0.6cm}
\end{table}
(\textit{i}) \textit{shortcut samples}, which induce sharp entropy drops and risk premature policy collapse, and 
(\textit{ii}) \textit{noisy samples} with near-zero covariance that contribute little to entropy change and slow learning. 
These findings motivate our dynamically adjusted clipping interval $[\ell, u]$, whose bounds are adaptively set per batch as the $5$th and $90$th percentiles of $\{|c(s_t,a_t)|\}$. 
We further analyze the sensitivity to clipping ratios through a hyperparameter study, as shown in Fig.~\ref{fig:ablation}.
\textbf{Analysis of Clipped Samples.}
We analyze the distribution of clipped samples to understand how entropy-guided selection shapes training. 
Fig.~\ref{fig:touch_histogram} shows accumulated retained (lines) and clipped (histograms) samples over time. 
Most clipped samples originate from human interventions. Many removed transitions lie outside the effective manipulation region, indicating limited gradient contribution, while others within the workspace correspond to redundant states with minimal entropy impact. 
Clipping is determined by entropy influence rather than spatial heuristics; consistent with ablation results, simple spatial filtering does not achieve comparable performance. Notably, shortcut and noisy samples exist inside the workspace and are selectively removed. 
Overall, entropy-bounded clipping mitigates shortcut-induced collapse and suppresses low-informative corrections, leading to stable learning.

\vspace{-0.3cm}
\subsection{Ablation Studies and Hyperparameter Analysis}
\label{sec:ablation}

\textbf{Ablation Study.} 
We conduct mechanism-isolation experiments on the \textit{Insert Block} task to examine whether the performance gains arise from entropy reasoning rather than generic regularization. 
Specifically, we compare E2HiL with matched-rate Random Drop, Advantage Clipping, Spatial Clipping, and Prioritized Experience Replay (PER). 
As shown in Fig.~\ref{fig:ablation}, none of these variants achieve comparable improvements, indicating that the gains stem from entropy-informed sample selection instead of simple data reduction, gradient magnitude clipping, or heuristic entropy control.

\textbf{Hyperparameter Analysis.} 
We evaluate sensitivity to clipping bounds (5–90, 10–90, 5–80) and Monte Carlo sample counts $K \in \{1,4,8,16\}$ (Fig.~\ref{fig:ablation}). 
Across reasonable ranges, performance remains stable, demonstrating that E2HiL is robust to entropy threshold and estimator configurations. 
Using $K=1$ leads to larger fluctuations, while $K=8$ (ours) and $K=16$ achieve nearly identical results, indicating diminishing returns beyond moderate sampling. Although the influence estimator depends on critic accuracy, its effect diminishes as value estimates stabilize during training. 
We further validate robustness using critic calibration (Cal-QL) and ensemble critics (Table~\ref{tab:Q_variant_ablation}), where consistent improvements confirm that E2HiL does not rely on accurate Q estimates in early phases. Moreover, entropy-guided sample selection stabilizes value updates by suppressing high-variance transitions, enabling stable Q learning; in practice, policy and critic improve in a co-evolving manner rather than depending on a perfectly one at initialization.


\vspace{0.2cm}
\section{Conclusion}
\label{sec:conclusion}

In this paper, we presented \method, an efficient real-world human-in-the-loop RL framework with entropy-guided sample selection. By modeling sample-induced entropy dynamics via covariance-based influence functions, \method filters shortcut and noisy samples to maintain stable entropy evolution and improve sample efficiency. 
Extensive real-world experiments across diverse tasks demonstrate that \method consistently achieves higher success rates with fewer human interventions than state-of-the-art baselines. Importantly, the method is policy-agnostic and embodiment-agnostic, serving as a plug-and-play optimization module across different learning frameworks and robotic platforms while preserving a favorable exploration–performance trade-off.
In future work, we plan to extend \method to scalable multi-task real-world RL for VLA models and further improve covariance estimation with more robust value-learning techniques.


\bibliographystyle{IEEEtran} 
\bibliography{sec/reference}

@STRING{icra = {Proceedings of the IEEE International Conference on Robotics and Automation (ICRA)}}

@STRING{iros = {Proceedings of the IEEE/RSJ International Conference on Intelligent Robots and Systems (IROS)}}

@STRING{nips = {Proceedings of Advances in Neural Information Processing Systems}}

@STRING{icml = {Proceedings of the International Conference on Machine Learning (ICML)}}

@STRING{ijrr = {International Journal of Robotics Research (IJRR)}}

@STRING{rss = {Proceedings of Robotics: Science and Systems (RSS)}}

@STRING{corl = {Conference on Robot Learning (CoRL)}}

@STRING{eccv = {European Conference on Computer Vision (ECCV)}}

@STRING{aaai = {Proceedings of the AAAI Conference on Artificial Intelligence (AAAI)}}

@article{luo2024precise,
  title={Precise and dexterous robotic manipulation via human-in-the-loop reinforcement learning},
  author={Luo, J and Xu, C and Wu, J and Levine, S},
  journal = {Science Robotics},
  year={2025},
  pages={1--54},
  doi = {10.1126/scirobotics.ads5033},
}

@inproceedings{renze2024effect,
  title={The effect of sampling temperature on problem solving in large language models},
  author={Renze, Matthew},
  booktitle={Findings of the association for computational linguistics: EMNLP 2024},
  pages={7346--7356},
  year={2024}
}

@article{zhang2024empowering,
  title={Empowering embodied manipulation: A bimanual-mobile robot manipulation dataset for household tasks},
  author={Zhang, Tianle and Li, Dongjiang and Li, Yihang and Zeng, Zecui and Zhao, Lin and Sun, Lei and Chen, Yue and Wei, Xuelong and Zhan, Yibing and Li, Lusong and others},
  journal={arXiv preprint arXiv:2405.18860},
  pages={1--24},
  year={2024}  
}

@article{abdolmaleki2018maximum,
  title={Maximum a posteriori policy optimisation},
  author={Abdolmaleki, Abbas and Springenberg, Jost Tobias and Tassa, Yuval and Munos, Remi and Heess, Nicolas and Riedmiller, Martin},
  journal={arXiv preprint arXiv:1806.06920},
  year={2018}
}

@article{eysenbach2021maximum,
  title={Maximum entropy RL (provably) solves some robust RL problems},
  author={Eysenbach, Benjamin and Levine, Sergey},
  journal={arXiv preprint arXiv:2103.06257},
  year={2021}
}

@article{deng2025survey,
  title={A Survey on Reinforcement Learning of Vision-Language-Action Models for Robotic Manipulation},
  author={Deng, Haoyuan and Wu, Zhenyu and Liu, Haichao and Guo, Wenkai and Xue, Yuquan and Shan, Ziyu and Zhang, Chuanrui and Jia, Bofang and Ling, Yuan and Lu, Guanxing and others},
  journal={Authorea Preprints},
  publisher={Authorea},
  year={2025}
}

@article{lu2025vla,
  title={Vla-rl: Towards masterful and general robotic manipulation with scalable reinforcement learning},
  author={Lu, Guanxing and Guo, Wenkai and Zhang, Chubin and Zhou, Yuheng and Jiang, Haonan and Gao, Zifeng and Tang, Yansong and Wang, Ziwei},
  journal={arXiv preprint arXiv:2505.18719},
  year={2025}
}

@article{chen2025conrft,
  title={Conrft: A reinforced fine-tuning method for vla models via consistency policy},
  author={Chen, Yuhui and Tian, Shuai and Liu, Shugao and Zhou, Yingting and Li, Haoran and Zhao, Dongbin},
  journal=rss,
  year={2025}
}

@article{hung2025nora,
  title={Nora-1.5: A vision-language-action model trained using world model-and action-based preference rewards},
  author={Hung, Chia-Yu and Majumder, Navonil and Deng, Haoyuan and Renhang, Liu and Ang, Yankang and Zadeh, Amir and Li, Chuan and Herremans, Dorien and Wang, Ziwei and Poria, Soujanya},
  journal={arXiv preprint arXiv:2511.14659},
  year={2025}
}

@article{guo2025vla,
  title={Vla-reasoner: Empowering vision-language-action models with reasoning via online monte carlo tree search},
  author={Guo, Wenkai and Lu, Guanxing and Deng, Haoyuan and Wu, Zhenyu and Tang, Yansong and Wang, Ziwei},
  journal=icra,
  year={2026}
}

@article{torne2024reconciling,
  title={Reconciling reality through simulation: A real-to-sim-to-real approach for robust manipulation},
  author={Torne, Marcel and Simeonov, Anthony and Li, Zechu and Chan, April and Chen, Tao and Gupta, Abhishek and Agrawal, Pulkit},
  journal={arXiv preprint arXiv:2403.03949},
  pages={1--23},
  year={2024}
}

@article{zhang2021reinforcement,
  title={Reinforcement learning for robot research: A comprehensive review and open issues},
  author={Zhang, Tengteng and Mo, Hongwei},
  journal={International Journal of Advanced Robotic Systems},
  year={2021},
}

@article{ju2022transferring,
  title={Transferring policy of deep reinforcement learning from simulation to reality for robotics},
  author={Ju, Hao and Juan, Rongshun and Gomez, Randy and Nakamura, Keisuke and Li, Guangliang},
  journal={Nature Machine Intelligence},
  volume={4},
  pages={1077--1087},
  year={2022},
  publisher={Nature Publishing Group UK London}
}

@inproceedings{tang2025deep,
  title={Deep reinforcement learning for robotics: A survey of real-world successes},
  author={Tang, Chen and Abbatematteo, Ben and Hu, Jiaheng and Chandra, Rohan and Mart{\'\i}n-Mart{\'\i}n, Roberto and Stone, Peter},
  booktitle=aaai,
  volume={39},
  number={27},
  pages={28694--28698},
  year={2025}
}

@article{spencer2022expert,
  title={Expert intervention learning: An online framework for robot learning from explicit and implicit human feedback},
  author={Spencer, Jonathan and Choudhury, Sanjiban and Barnes, Matthew and Schmittle, Matthew and Chiang, Mung and Ramadge, Peter and Srinivasa, Sidd},
  journal={Autonomous Robots},
  volume={46},
  number={1},
  pages={99--113},
  year={2022},
  publisher={Springer}
}

@inproceedings{kelly2019hg,
  title={Hg-dagger: Interactive imitation learning with human experts},
  author={Kelly, Michael and Sidrane, Chelsea and Driggs-Campbell, Katherine and Kochenderfer, Mykel J},
  booktitle={2019 International Conference on Robotics and Automation (ICRA)},
  pages={8077--8083},
  year={2019},
  organization={IEEE}
}

@inproceedings{lee2020guided,
  title={Guided uncertainty-aware policy optimization: Combining learning and model-based strategies for sample-efficient policy learning},
  author={Lee, Michelle A and Florensa, Carlos and Tremblay, Jonathan and Ratliff, Nathan and Garg, Animesh and Ramos, Fabio and Fox, Dieter},
  booktitle={2020 IEEE international conference on robotics and automation (ICRA)},
  pages={7505--7512},
  year={2020}
}

@inproceedings{clavera2018model,
  title={Model-based reinforcement learning via meta-policy optimization},
  author={Clavera, Ignasi and Rothfuss, Jonas and Schulman, John and Fujita, Yasuhiro and Asfour, Tamim and Abbeel, Pieter},
  booktitle=corl,
  pages={617--629},
  year={2018}
}

@article{pinosky2023hybrid,
  title={Hybrid control for combining model-based and model-free reinforcement learning},
  author={Pinosky, Allison and Abraham, Ian and Broad, Alexander and Argall, Brenna and Murphey, Todd D},
  journal=ijrr,
  volume={42},
  number={6},
  pages={337--355},
  year={2023},
  publisher={SAGE Publications Sage UK: London, England}
}

@inproceedings{zhong2024empowering,
  title={Empowering embodied visual tracking with visual foundation models and offline rl},
  author={Zhong, Fangwei and Wu, Kui and Ci, Hai and Wang, Churan and Chen, Hao},
  booktitle=eccv,
  pages={139--155},
  year={2024},
  organization={Springer}
}

@article{kawaharazuka2024real,
  title={Real-world robot applications of foundation models: A review},
  author={Kawaharazuka, Kento and Matsushima, Tatsuya and Gambardella, Andrew and Guo, Jiaxian and Paxton, Chris and Zeng, Andy},
  journal={Advanced Robotics},
  volume={38},
  number={18},
  pages={1232--1254},
  year={2024},
  publisher={Taylor \& Francis}
}

@inproceedings{memarian2021self,
  title={Self-supervised online reward shaping in sparse-reward environments},
  author={Memarian, Farzan and Goo, Wonjoon and Lioutikov, Rudolf and Niekum, Scott and Topcu, Ufuk},
  booktitle={2021 IEEE/RSJ International Conference on Intelligent Robots and Systems (IROS)},
  pages={2369--2375},
  year={2021},
  organization={IEEE}
}

@article{cao2024survey,
  title={Survey on large language model-enhanced reinforcement learning: Concept, taxonomy, and methods},
  author={Cao, Yuji and Zhao, Huan and Cheng, Yuheng and Shu, Ting and Chen, Yue and Liu, Guolong and Liang, Gaoqi and Zhao, Junhua and Yan, Jinyue and Li, Yun},
  journal={IEEE Transactions on Neural Networks and Learning Systems},
  year={2024},
  publisher={IEEE}
}

@inproceedings{gupta2021reset,
  title={Reset-free reinforcement learning via multi-task learning: Learning dexterous manipulation behaviors without human intervention},
  author={Gupta, Abhishek and Yu, Justin and Zhao, Tony Z and Kumar, Vikash and Rovinsky, Aaron and Xu, Kelvin and Devlin, Thomas and Levine, Sergey},
  booktitle={2021 IEEE International Conference on Robotics and Automation (ICRA)},
  pages={6664--6671},
  year={2021},
  organization={IEEE}
}

@article{yang2024reset,
  title={Reset-free Reinforcement Learning with World Models},
  author={Yang, Zhao and Moerland, Thomas M and Preuss, Mike and Plaat, Aske and Hu, Edward S},
  journal={arXiv preprint arXiv:2408.09807},
  year={2024}
}

@article{dromnelle2023reducing,
  title={Reducing computational cost during robot navigation and human--robot interaction with a human-inspired reinforcement learning architecture},
  author={Dromnelle, R{\'e}mi and Renaudo, Erwan and Chetouani, Mohamed and Maragos, Petros and Chatila, Raja and Girard, Beno{\^\i}t and Khamassi, Mehdi},
  journal={International Journal of Social Robotics},
  volume={15},
  number={8},
  pages={1297--1323},
  year={2023},
  publisher={Springer}
}

@article{cui2025entropy,
  title={The entropy mechanism of reinforcement learning for reasoning language models},
  author={Cui, Ganqu and Zhang, Yuchen and Chen, Jiacheng and Yuan, Lifan and Wang, Zhi and Zuo, Yuxin and Li, Haozhan and Fan, Yuchen and Chen, Huayu and Chen, Weize and others},
  journal={arXiv preprint arXiv:2505.22617},
  year={2025}
}

@article{cheng2025reasoning,
  title={Reasoning with exploration: An entropy perspective},
  author={Cheng, Daixuan and Huang, Shaohan and Zhu, Xuekai and Dai, Bo and Zhao, Wayne Xin and Zhang, Zhenliang and Wei, Furu},
  journal={arXiv preprint arXiv:2506.14758},
  year={2025}
}

@inproceedings{tiapkin2023fast,
  title={Fast rates for maximum entropy exploration},
  author={Tiapkin, Daniil and Belomestny, Denis and Calandriello, Daniele and Moulines, Eric and Munos, Remi and Naumov, Alexey and Perrault, Pierre and Tang, Yunhao and Valko, Michal and Menard, Pierre},
  booktitle=icml,
  pages={34161--34221},
  year={2023},
  organization={PMLR}
}

@article{li2025simplevla,
  title={SimpleVLA-RL: Scaling VLA Training via Reinforcement Learning},
  author={Li, Haozhan and Zuo, Yuxin and Yu, Jiale and Zhang, Yuhao and Yang, Zhaohui and Zhang, Kaiyan and Zhu, Xuekai and Zhang, Yuchen and Chen, Tianxing and Cui, Ganqu and others},
  journal={arXiv preprint arXiv:2509.09674},
  year={2025}
}

@InProceedings{pmlr-v80-haarnoja18b,
  title = 	 {Soft Actor-Critic: Off-Policy Maximum Entropy Deep Reinforcement Learning with a Stochastic Actor},
  author =       {Haarnoja, Tuomas and Zhou, Aurick and Abbeel, Pieter and Levine, Sergey},
  pages = 	 {1861--1870},
  year = 	 {2018},
  volume = 	 {80},
  series = 	 {Proceedings of Machine Learning Research},
  publisher =    {PMLR},
}

@article{chen2024rlingua,
  title={Rlingua: Improving reinforcement learning sample efficiency in robotic manipulations with large language models},
  author={Chen, Liangliang and Lei, Yutian and Jin, Shiyu and Zhang, Ying and Zhang, Liangjun},
  journal={IEEE Robotics and Automation Letters},
  volume={9},
  number={7},
  pages={6075--6082},
  year={2024},
  publisher={IEEE}
}

@article{lei2025rl,
  title={RL-100: Performant Robotic Manipulation with Real-World Reinforcement Learning},
  author={Lei, Kun and Li, Huanyu and Yu, Dongjie and Wei, Zhenyu and Guo, Lingxiao and Jiang, Zhennan and Wang, Ziyu and Liang, Shiyu and Xu, Huazhe},
  journal={arXiv preprint arXiv:2510.14830},
  year={2025}
}

@article{yang2023sim,
  title={Sim-to-real model-based and model-free deep reinforcement learning for tactile pushing},
  author={Yang, Max and Lin, Yijiong and Church, Alex and Lloyd, John and Zhang, Dandan and Barton, David AW and Lepora, Nathan F},
  journal={IEEE Robotics and Automation Letters},
  volume={8},
  number={9},
  pages={5480--5487},
  year={2023},
  publisher={IEEE}
}

@inproceedings{ball2023efficientonlinereinforcementlearning,
  title={Efficient online reinforcement learning with offline data},
  author={Ball, Philip J and Smith, Laura and Kostrikov, Ilya and Levine, Sergey},
  booktitle=icml,
  pages={1577--1594},
  year={2023},
  organization={PMLR}
}

@article{nakamoto2024calqlcalibratedofflinerl,
  title={Cal-ql: Calibrated offline rl pre-training for efficient online fine-tuning},
  author={Nakamoto, Mitsuhiko and Zhai, Simon and Singh, Anikait and Sobol Mark, Max and Ma, Yi and Finn, Chelsea and Kumar, Aviral and Levine, Sergey},
  journal=NIPS,
  volume={36},
  pages={62244--62269},
  year={2023}
}

@article{muraleedharan2025selective,
  title={Selective Progress-Aware Querying for Human-in-the-Loop Reinforcement Learning},
  author={Muraleedharan, Anujith and others},
  journal={arXiv preprint arXiv:2509.20541},
  year={2025}
}

\end{document}